\documentclass{svproc}
\usepackage{url}

\usepackage{amsmath}
\usepackage{amsfonts}
\usepackage{subfigure}
\usepackage{float}
\usepackage{graphicx}
\usepackage{cite}

\begin{document}
\mainmatter              
\title{An Extended Multi-Model Regression Approach for Compressive Strength  Prediction and Optimization of a Concrete Mixture}
\titlerunning{Multi-Model Regression for Concrete CS Prediction and Mixture Opt.}  
%
\author{Seyed Arman Taghizadeh Motlagh\inst{1} \and Mehran Naghizadehrokni\inst{2}}
\authorrunning{Arman Taghizadeh et al.} 
%
%
\institute{$^{1}$Azad University, Central Tehran Branch (IAUCTB),\\ $^{2}$RWTH Aachen University, Lehrstuhl für Geotechnik im Bauwesen und
	Institut für Geomechanik und Untergrundtechnik}

\maketitle              

\begin{abstract}
Due to the significant delay and cost associated with experimental tests, a model based evaluation of concrete compressive strength is of high value, both for the purpose of strength prediction as well as the mixture optimization. In this regard, several recent studies have employed state-of-the-art regression models in order to achieve a good prediction model, employing available experimental data sets. Nevertheless, while each of the employed models can better adapt to a specific nature of the input data, the accuracy of each individual model is limited due to the sensitivity to the choice of hyperparameters and the learning strategy. In the present work, we take a further step towards improving the accuracy of the prediction model via the weighted combination of multiple regression methods. Moreover, a heuristic Genetic Algorithm (GA)-based multi-objective mixture optimization is proposed, building on the obtained multi-regression model. In particular, we present a data-aided framework where the regression methods based on artificial neural network (ANN), random forest regression, and polynomial regression are jointly implemented to predict the compressive strength of concrete. The outcome of the individual regression models are then combined via a linear weighting strategy and optimized over the training data set as a quadratic convex optimization problem. It is worth mentioning that due to the convexity of the formulated problem, the globally optimum weighting strategy is obtained via standard numerical solvers. Employing the proposed GA-based optimization, a Pareto front of the cost-CS trade-off has been obtained employing the available data set. Moreover, the resulting accuracy of the proposed multi-model prediction method is shown to outperform the available single-model regression methods in the literature by a valuable margin, via numerical simulations.
\end{abstract}

\section{INTRODUCTION}   
The compressive strength (CS) of concrete, as a key element in many instances of construction industry, is known as an important metric for performance evaluation. While the required CS level vary depending on the usecase and the target quality, CS is highly dependent on the choice of the material and mixture parameters~\cite{Ref_1}. In this regard, while experimental evaluation of a specific mixture strategy is both time consuming and expensive, the need for a model-based prediction of CS level for a known mixture appears as a viable alternative~\cite{Ref_2, Ref_3}. Furthermore, obtaining an accurate predictive model enables an optimization of mixture strategy for a specific material budget limit or CS level target, see, e.g.,~\cite{Ref_4, Ref_7} for similar applications. In this regard, several recent studies have employed state-of-the-art regression models in order to achieve a good prediction, employing available experimental data sets. A common shortcoming of the presented model-based approaches is the shortage of sufficient data set that are reported with sufficient details, and can be used for the purpose of model training and regression analysis. In this regard, a collection of $17$ separately reported laboratory data sets have been combined and validated by \cite{Ref_2}, leading to a data set with $1000$ experimental results. An artificial neural network (ANN)-based regression model is then employed for modeling and validating the model accuracy employing the combined data sets~\cite{Ref_2, Ref_3, Ref_10}. 

The data-aided machine learning regression methods have recently received a growing attention for various engineering problems, both due to their good performance and little restrictive assumptions on the nature of the input data~\cite{Ref_13}. In addition to the widely-used ANN-based models, the predictive model for CS level prediction of concrete can be also studied using other regression models, e.g., using support vector regression~\cite{Ref_5, Ref_6, Ref_7}, decision tree and random forest regression and $K$-nearest neighbors regression~\cite{Ref_7, Ref_8, Ref_9}. Please note that the obtained regression models can be used both for the purpose of low-cost CS prediction as well as the model-based mixture optimization~\cite{Ref_A, Ref_B}. The latter use-case becomes more pronounced when considering multiple simultaneous objectives and/or design constraints together with the high cost of the experimental tests, which renders the try-and-error experimental designs as impractical. In this regard, heuristic optimization methods based on Genetic-Algorithm (GA)~\cite{Ref_A, Ref_B, Ref_D} and Particle Swarm Optimization (PSO) methods~\cite{Ref_C} have been proposed, utilizing the developed regression models for mixture CS prediction as a fitness function. Nevertheless, while each of the employed regression models can better adapt to a specific nature of the input data, the accuracy of each individual model is limited for all of the parameters range due to the sensitivity to the choice of hyperparameters and the learning strategy. In this regard, the work in~\cite{Ref_7} have reported the performance of multiple individual regression models for predicting CS of concrete, where ANN is found to be the best performing model and later used for mixture optimization. To the best of authors knowledge, a study on the multi-model prediction for CS of concrete is still an open problem.   

In the present work, we take a further step towards improving the accuracy of the CS prediction model via the weighted combination of multiple regression methods. In particular, we propose a data-aided framework where the regression methods based on ANN, random forest regression, and polynomial regression are jointly implemented to predict the CS of the ready-mix concrete. The outcome of the individual regression models are then combined via a linear weighting strategy, where the weights are optimized over the training data set to minimize the mean squared error (MSE) of the prediction via a quadratic convex optimization problem. It is worth mentioning that due to the convexity of the formulated problem, the globally optimum weighting strategy can obtained via standard numerical solvers. Furthermore, a heuristic Genetic Algorithm (GA)-based multi-objective mixture optimization is proposed, building on the obtained multi-regression model. In particular, we use the multi-objective variant of the GA optimization, namely, the Non-dominated Sorting Genetic Algorithm II (NSGA-II), in order to obtain a Pareto front of the cost-CS trade-off employing the available data set. Numerical simulations indicate the reduction of the resulting prediction MSE over the test data set by $13\%$ over the best individual regression model, indicating a significant gain via the proposed multi-model approach compared to the traditional single-model regression approaches.      

The rest of this paper is organized as follows. In Section~\ref{sec_data}, the nature of the data structure and the employed dataset is explained. The problem of multi-variable regression is mathematically formulated in Section~\ref{sec_problemformulation}, which presents the main goal of this paper. In Section~\ref{sec_multimodel} the implemented multi-model regression method and weighting optimization is presented. Building on the obtained regression model, a heuristic NSGA-II-based optimization method is presented in Section~\ref{sec_optimize}. The results of the numerical evaluations are reported in Section~\ref{sec_results}. We conclude this paper in Section~\ref{sec_conclude} by summarizing the main findings.  

\section{MIXTURE PARAMETERS AND DATA SET} \label{sec_data}
In this work, we use the collected experimental data set reported by \cite{Ref_2}, where each data point represents the mixture parameters, including: \textbf{\textit{cement, fly ash, blast furnace slag, water, superplasticizer, coarse aggregate, fine aggregate, age of curing}} as the independent variables and the resulting concrete CS, i.e., \textbf{\textit{$f_c$}} as the dependent variable. In particular, the data set with the total size of $1030$ is randomly divided into a training data set of size $721$, ($70\%$ of the available data set) which is used for the purpose of model training, and  the test data set with the size $309$ ($30\%$ of the available data set), used for evaluating the prediction accuracy. Please see Table~1 for a detailed parameter definition. 
\begin{table}[h]
	\begin{center}
		\begin{tabular}{*{5}{c}}
			\hline
			\textbf{Variable} & \textbf{Min.} & \textbf{Max.} & \textbf{Average} & \textbf{SD} \\
			\hline
			\textbf{Cement (kg/m3):} & $102$  & $540$ & $281.2$ & $104.5$ \\
			\textbf{Fly Ash (kg/m3):} & $0$  & $359.4$ & $73.9$ & $86.2$ \\
			\textbf{Blast furnace Slag (kg/m3):} & $0$  & $200.1$ & $54.2$ & $64$ \\
			\textbf{Water (kg/m3):} & $247$  & $121.75$ & $181.6$ & $21.3$ \\
			\textbf{Superplasticizer (kg/m3):} & $0$  & $32$ & $6.2$ & $6$ \\
			\textbf{Coarse Aggregate (kg/m3):} & $801$  & $1145$ & $972.9$ & $77.7$ \\
			\textbf{Fine Aggregate (kg/m3):} & $992.6$  & $594$ & $773.6$ & $80.1$ \\
			\textbf{Age of Curing (Day):} & $1$  & $365$ & $91.2$ & $63.1$ \\
			\textbf{fc (MPa):} & $2.33$  & $82.6$ & $35.8$ & $16.7$ \\
			\hline   
		\end{tabular}
	\end{center}
	\caption{Parameter definitions, including the range, average and Standard Deviation (SD) of the input parameters.}
\end{table}  
\subsection{Data Set Analysis}
\begin{figure*}[!ht]
	\subfigure[Age of curing (Day)]{\includegraphics[height = 4cm, width = 6cm]{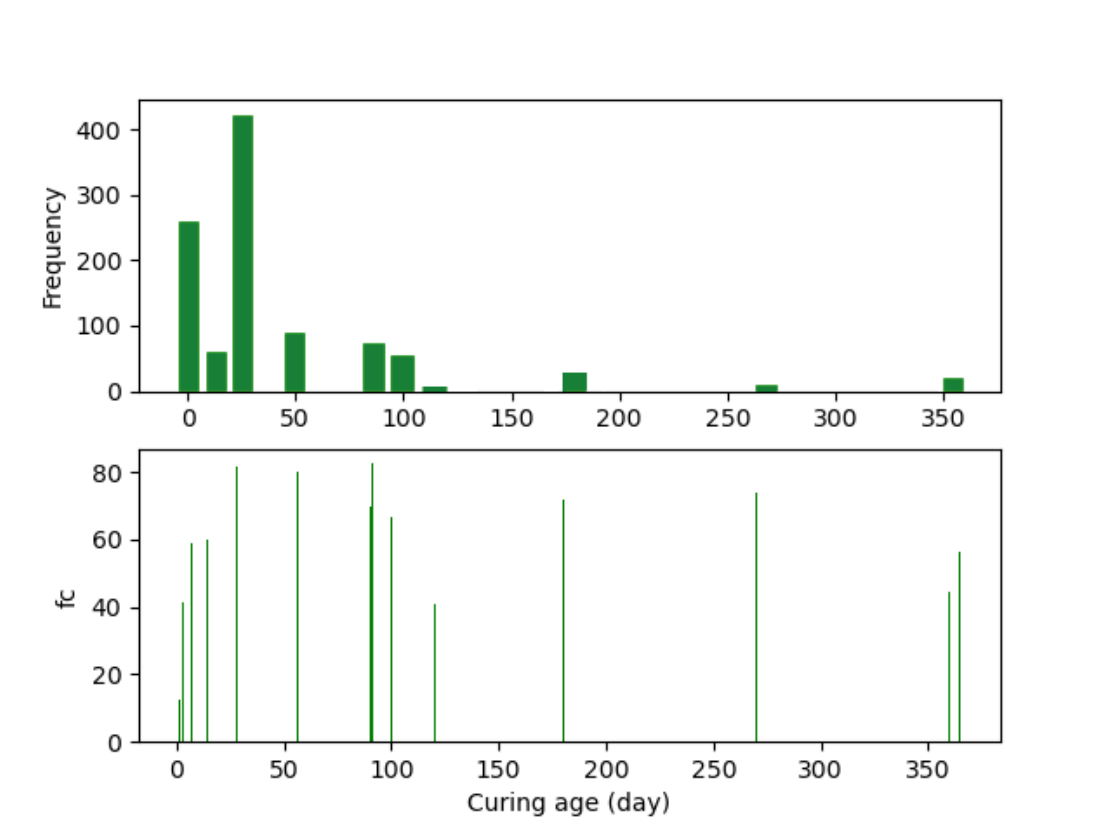} \label{fig_age} }
	\subfigure[Cement (kg/$m^{3})$]{\includegraphics[height = 4cm, width = 6cm]{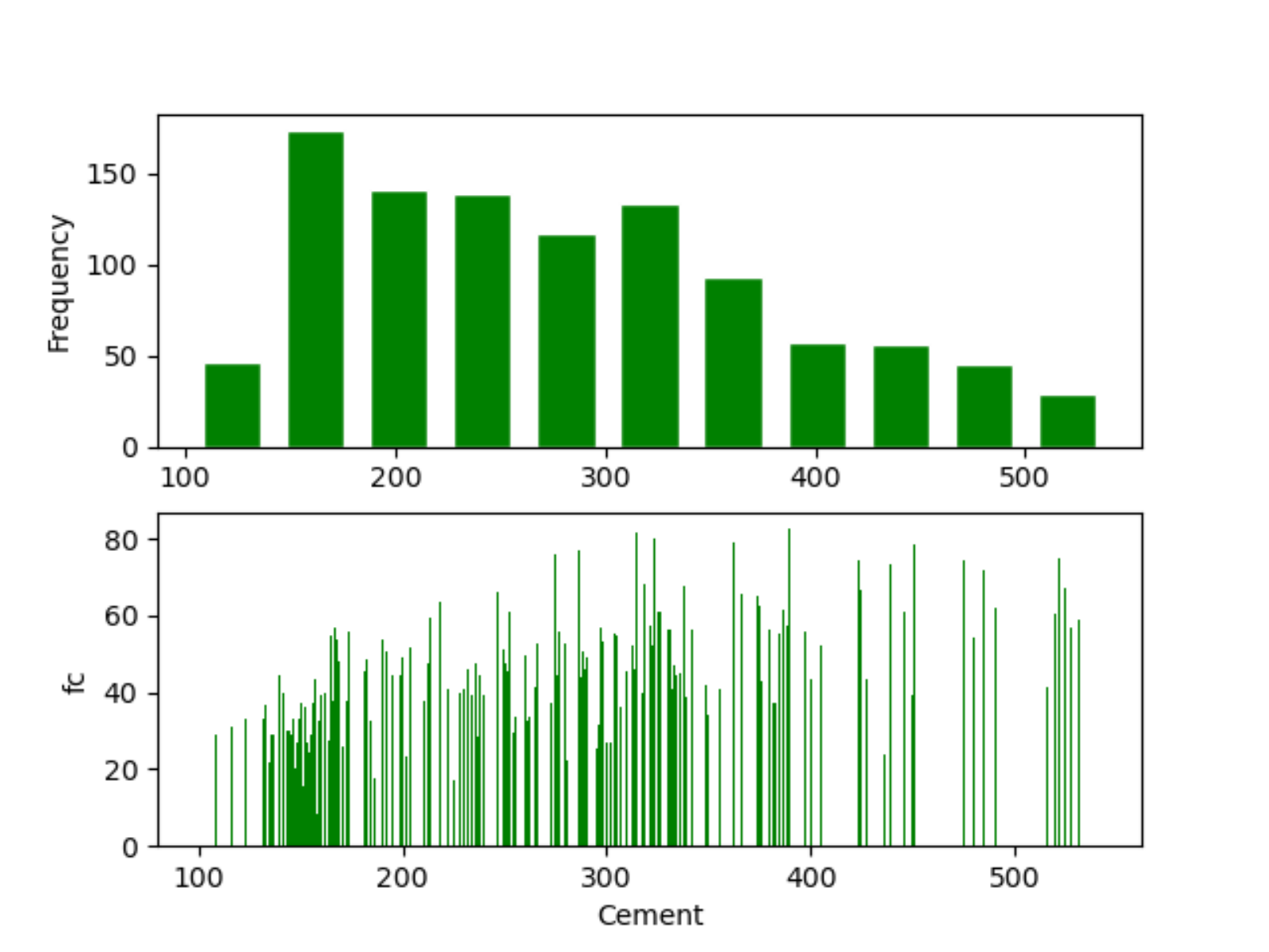} \label{fig_cement} }
	\subfigure[Water (kg/$m^{3})$]{\includegraphics[height = 4cm, width = 6cm]{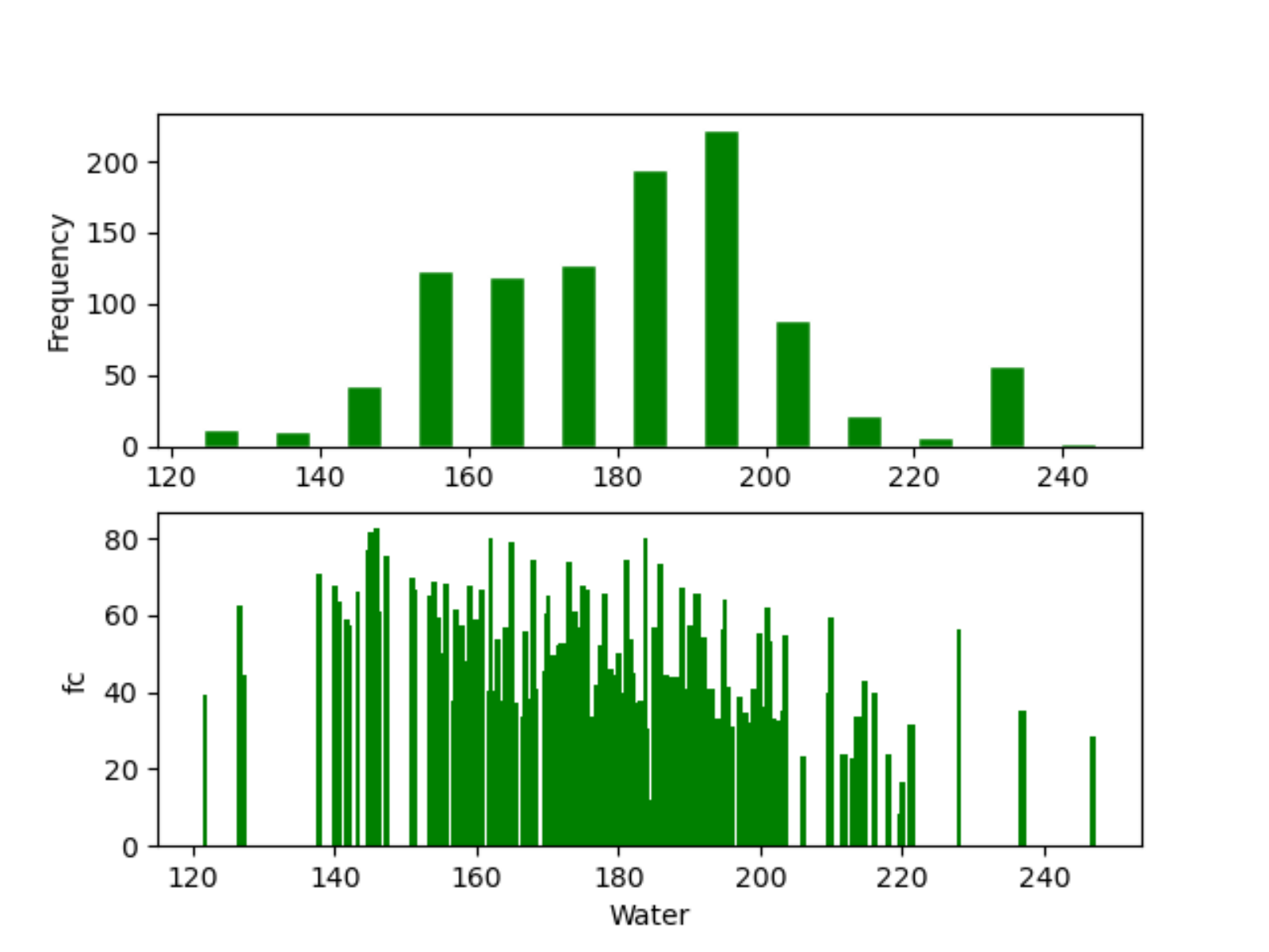} \label{fig_cement} }
	\subfigure[Superplasticizer (kg/$m^{3})$]{\includegraphics[height = 4cm, width = 6cm]{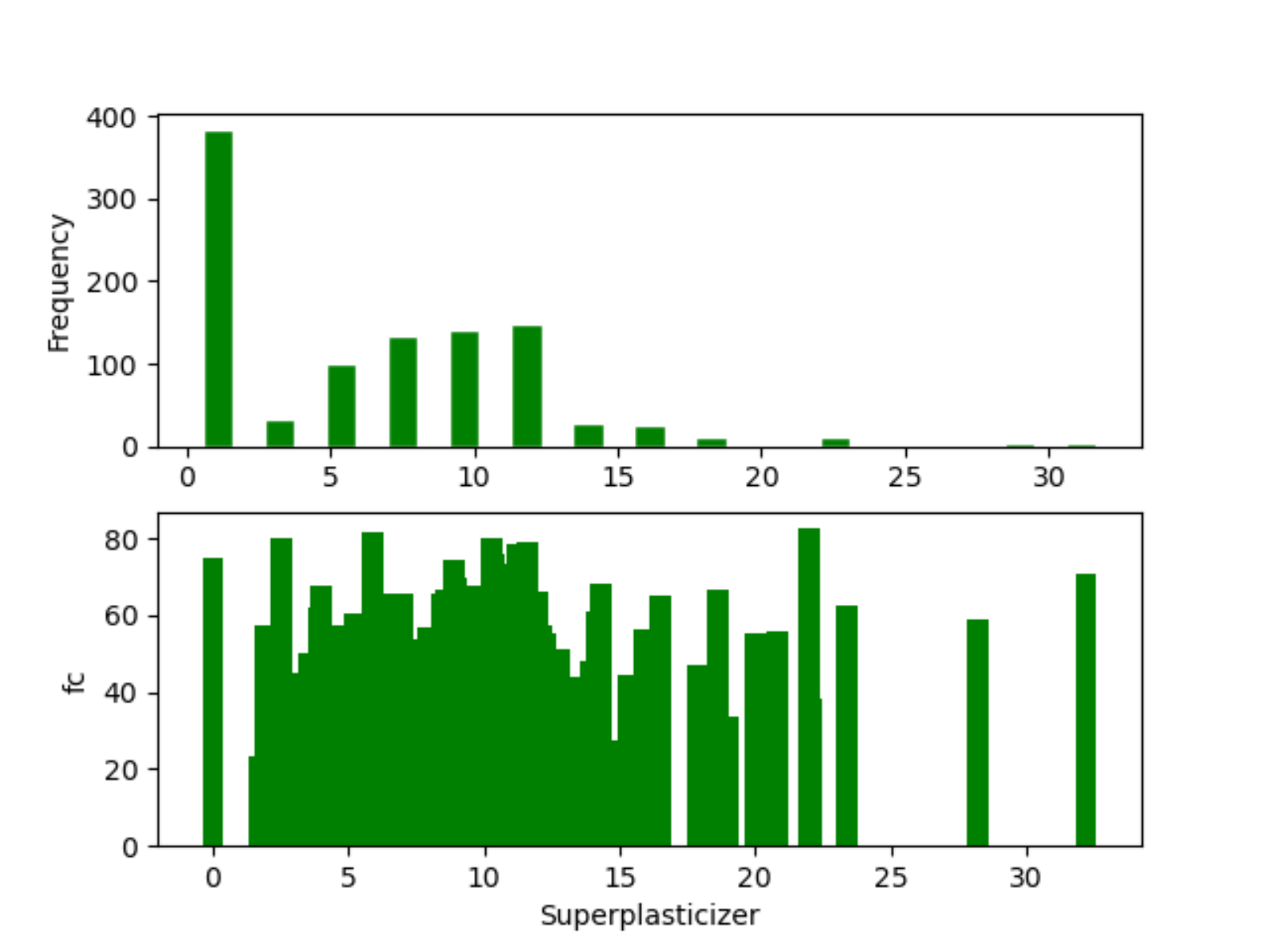} \label{fig_super} }
	\subfigure[Fine aggregate (kg/$m^{3})$]{\includegraphics[height = 4cm, width = 6cm]{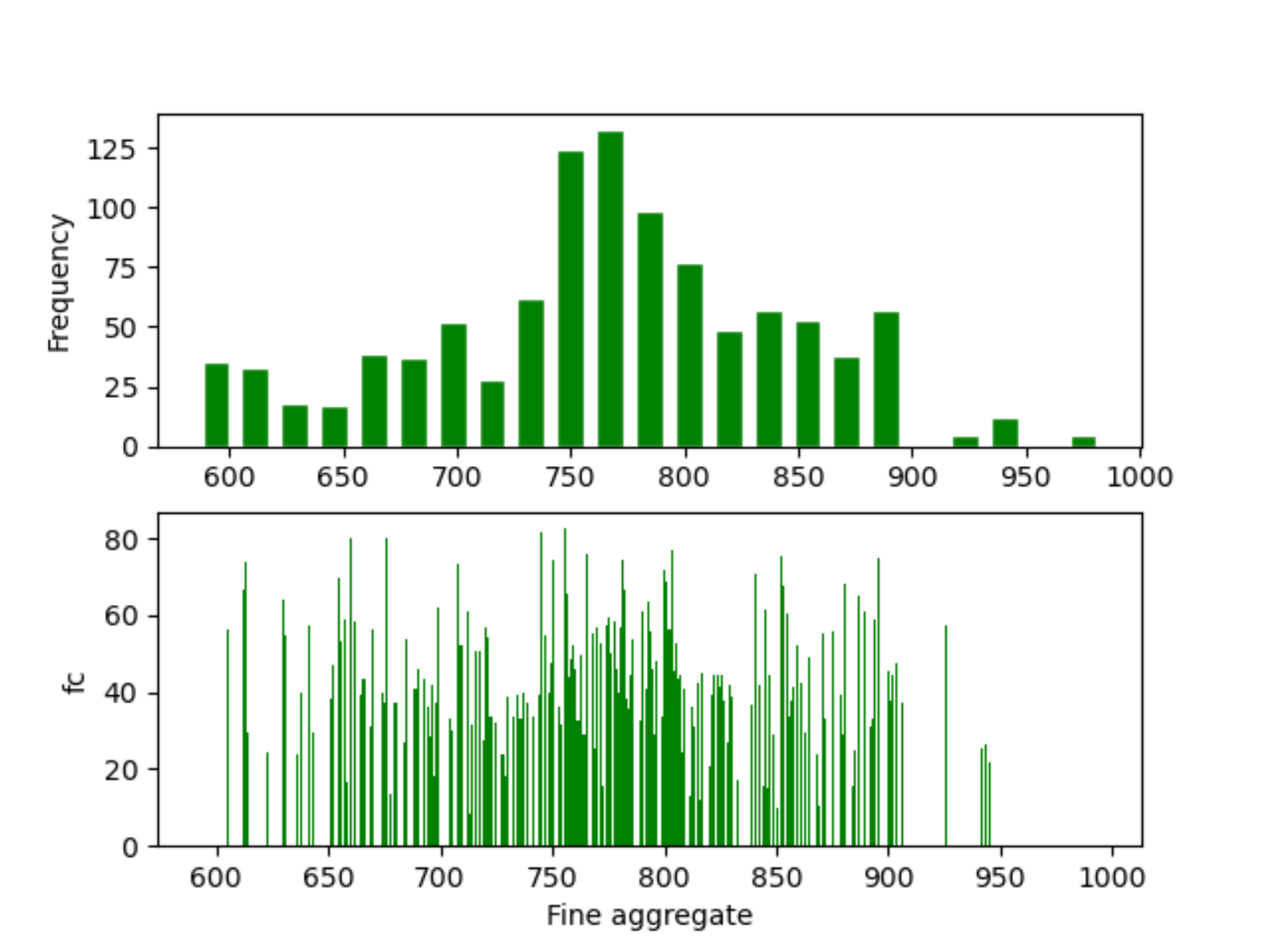} \label{fig_finegg} }
		\subfigure[Coarse aggregate (kg/$m^{3})$]{\includegraphics[height = 4cm, width = 6cm]{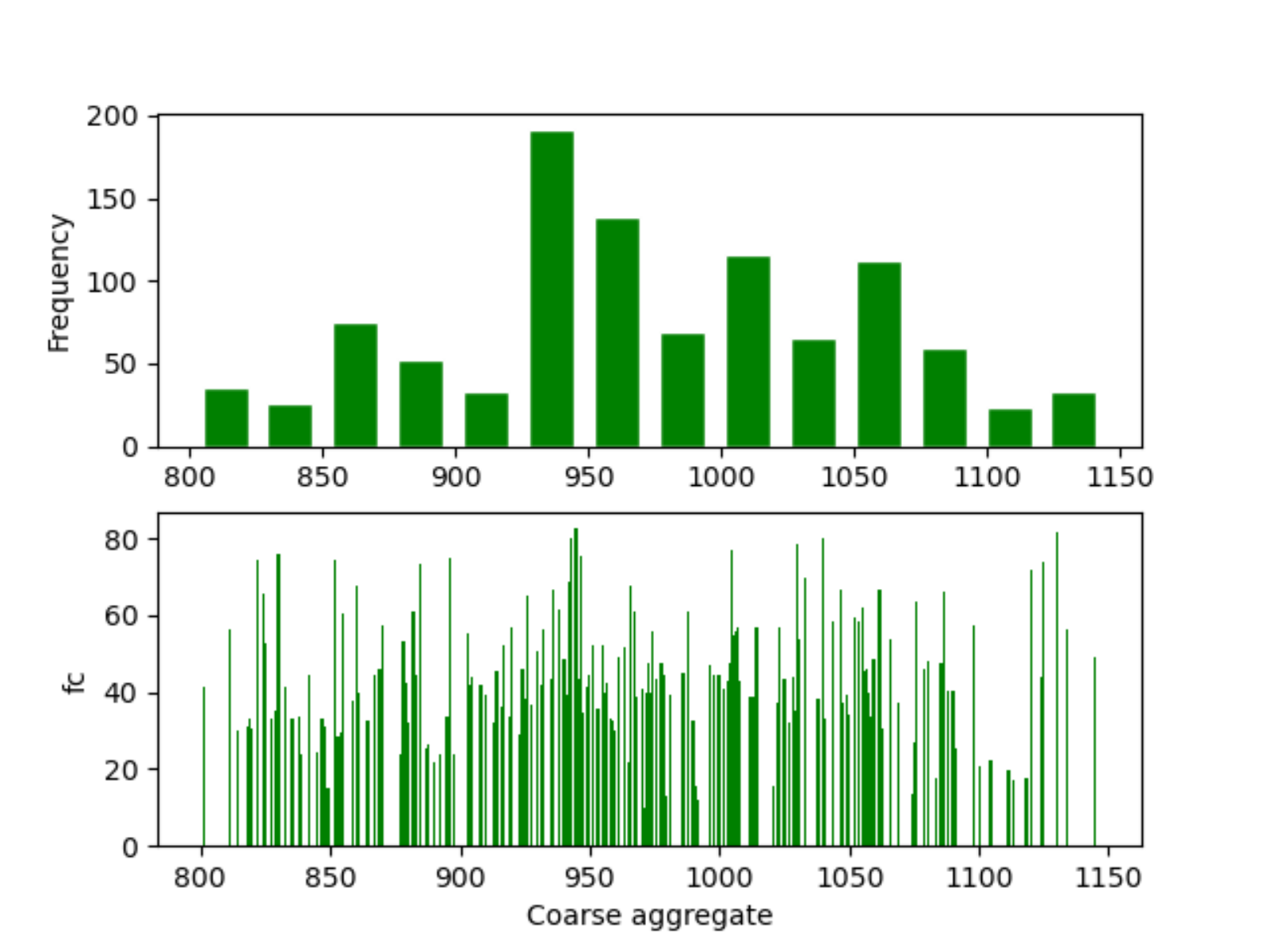} \label{fig_coagg} } 
		\caption{Distribution and the resulting compressive strength associated with the individual input parameters.} \label{fig_data}
\end{figure*}

	In the following we summarize useful observations on the type, value range and the statistical distribution of the parameters in the employed data set~\cite{Ref_2}. It is observed that the the parameter \textit{age of curing} is of discrete nature with distribution with a high frequency in the range $[1 \cdots 28]$ days. Moreover, the correlation between \textit{age of curing} and \textbf{\textit{$f_c$}} is apparent from Fig.~\ref{fig_data}(a). The parameter \textit{cement} follow a continuous nature with high frequency of occurrence in the range $[150 \cdots 200]\;kg/{m^3}$ for which an average CS of $25\; MPa$ is observed, see Fig.~\ref{fig_data}(b). The share of \textit{water} approximately follows a normal distribution with a mean value of $190\; kg/{m^3}$, see Fig.~\ref{fig_data}(c). The parameter \textit{superplasticizer} follows a continuous value distribution with a high frequency in the range $[2 \cdots 12]\;kg/{m^3}$ for which the average CS of $55\; MPa$ is observed, see Fig.~\ref{fig_data}(d). Moreover, the parameters \textit{coarse aggregate} and \textit{fine aggregate} follow a continuous distribution, respectively with a high frequency in the range $[925 \cdots 1025] \; kg/{m^3}$ for \textit{coarse aggregate} and an approximately normal distribution around $760 \; kg/{m^3}$ for \textit{fine aggregate} , please see Fig.~\ref{fig_data}(e) and Fig.~\ref{fig_data}(f), for better illustration.
\begin{figure}[!ht]
	\begin{center}         
		\includegraphics[width=11cm]{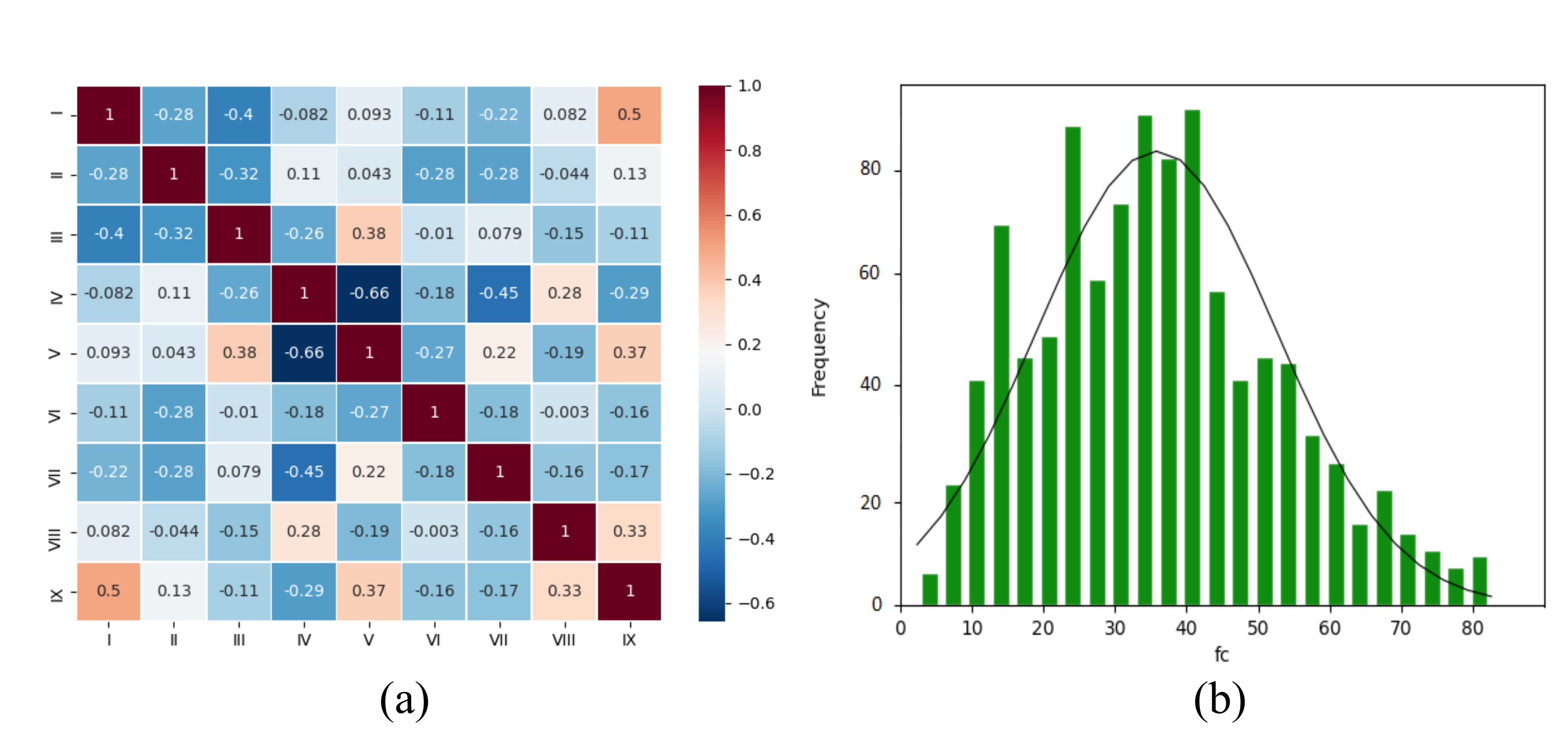}
		\caption{\textbf{\textit{(a)}} Evaluated Pierson coefficients. I: Cement, II: Blast furnace slag, III: Fly ash, IV: Water, V: Superplasticizer, VI: Coarse aggregate, VII: Fine aggregate, VIII: Age of curing, IX: $f_c$ \textbf{\textit{(b)}} Histogram and Distribution of $f_c$}
	\end{center}
\end{figure}


In order to ensure the validity of the chosen mixture parameters, we have evaluated the mutual dependence of the parameters by calculating the Pearson correlation coefficient for each parameter pair. As expected, it is observed that the mixture parameters have been chosen with little correlation, with the exception of the natural dependence among the superplasticizer and water, indicating the Pearson coefficient of $-0.66$. Furthermore, it is observed that the the linear dependence of the resulting $f_c$ to the individual parameters are negligible. For the detailed Pierson's correlation coefficients and the resulting $f_c$ statistics please see Fig.~2. In Fig.~2b, the histogram of the resulting $f_c$ is depicted over all reported parameter sets in the dataset. It is observed the resulting \textbf{\textit{$f_c$}} varies significantly depending on the choice of the mixture parameters. 

The insignificant linear dependence between $f_c$ and the individual mixture parameters from Fig.~2a, together with the high variance of the resulting $f_c$ over the reported mixture parameter set, indicate the significance of a non-linear multi-variable regression modeling for obtaining an accurate predictive model for $f_c$ given the mixture parameters.



\section{REGRESSION PROBLEM FORMULATION} \label{sec_problemformulation}
In this part, we provide a mathematical formulation of the intended problem as a general multi-variable regression. The provided formulation will be necessary to establish the proposed solution in Section~\ref{sec_multimodel}. In order to facilitate this, let us define the vector $\mathbf{x}_n \in \mathbb{R}^8$ and $y_n \in \mathbb{R},$ $\forall\; n  \in \mathcal{T}$, respectively representing the vector of mixture parameters and the resulting $f_c$ for the data instance $n$, where $\mathcal{T}$ ($\mathcal{T}^o$) denotes the training (test) set, such that $\mathcal{T}^o \cup \mathcal{T}$ builds the complete data set, see Section~\ref{sec_data} for dividing the used data set \cite{Ref_2} into training and test sets. This is the goal of the training phase to obtain a model that accurately predicts the value $y_n$ given $\mathbf{x}_n$, utilizing the available training data set, i.e., $n \in \mathcal{T}$. This is expressed as the following optimization problem 
\begin{align} \label{eq_1}
	\underset{\varphi(\cdot)}{\text{minimize}} \;\;\;\; \frac{1}{|\mathcal{T}|} \sum_{n \in \mathcal{T}} \; \left| \varphi(\mathbf{x}_n) - y_n \right|^2,
\end{align}
where $\varphi\; : \; \mathbb{R}^{8} \rightarrow \mathbb{R}$ is a multivariable and non-linear function, aiming to perform the regression task. Please note that the solution to the above problem varies depending on the used model, e.g., an ANN or a polynomial regression. In the following part we propose a mixed multi-model approach to solve the above problem with a high accuracy.

\section{MIXED MULTI-MODEL PREDICTIVE REGRESSION} \label{sec_multimodel}
We recall that each of the standard regression models, in particular the ANN~\cite{Ref_2, Ref_12}, random forest regression~\cite{Ref_7, Ref_12, Ref_14} and polynomial regression~\cite{Ref_15} yield a specific solution to the problem (\ref{eq_1}). Furthermore, it is known that depending on the nature of the input data statistics, each of the aforementioned models may outperform the others for specific structures of the input data~\cite{Ref_7, Ref_12}. In this regard, we propose a model utilizing a weighted combination of the solutions offered by the standard regression models. In the following, we first define each of the implemented standard regression models, reporting the utilized hyper-parameters and model architecture in each case. Afterwards, the multi-model regression strategy and weighting optimization is defined.
\subsection{ANN regression}
\begin{figure}[ht]
	\begin{center}
		\includegraphics[width=10cm, height=8cm]{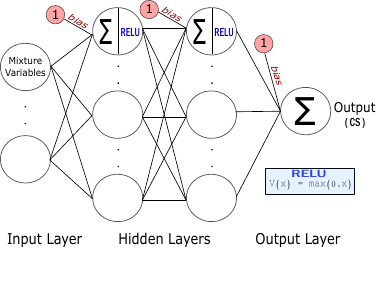} 
		\caption{ANN architecture} \label{fig_ANN}
	\end{center}
\end{figure}

ANNs, and in particular the deep ANNs have been recently presented as one of the successful methods for machine learning, due to their high generalization capability, ease of use and training, and fast inference~\cite{Ref_DeepL}. In particular, ANNs are capable of generating a large space of regression functions, by a parallel and multi-layered combination of a rather simple non-linear neuron, e.g., Sigmoid or ReLU functions~\cite{Ref_DeepL, Ref_7}. In this way, highly sophisticated regression functions can be synthesized by properly adjusting the weights and bias parameters at all layers. In order to implement this, let $V: \mathbb{R}\rightarrow \mathbb{R}$ represent the neuron function. The outcome of the $k$-th neuron at the layer $l$ and can be hence calculated as 
\begin{align}                          
	O_{l,k} = V \left( b_{lk} + \sum_{i} W_{{l-1},i,k}  O_{{l-1},i}  \right) 
\end{align}
where $b_{lk}$ represents the bias parameter and $W_{{l-1},i,k}$ represents the weight between the $l-1$-th and the $l$-th layer and from the $i$-th to the $k$-th neuron. For a general description of the implemented ANN, including the multi-layered structure and the role of bias and weight parameters please see Fig.~\ref{fig_ANN}. 

In the following, we summarize the implemented ANN architecture, the training protocol and the used hyperparameters.

\subsubsection{Network architecture}
We employ an ANN architecture with the input layer consisting of $8$ neurons, capturing the input mixture parameters, i.e., $\textbf{x}$, and a single neuron for the output layer, i.e., $f_c$, representing the predicted value of CS. Furthermore, we consider a network with five dense and fully-connected layers, where the ReLU activation function is used in all layers. The first to last hidden layers respectively include $64,64,32,16,16$ neurons. 

\subsubsection{ANN Training}  
The ANN training includes the process of tuning network parameters towards minimizing a loss function. In this regard, we have used the regularized mean squared error, defined as 
\begin{align}                  
	L\Big(\{ b_{lk} \}, \{W_{{l-1},i,k} \}\Big)  = \frac{1}{|\mathcal{T}|}\sum_{n \in \mathcal{T}} \; \left| \tilde{y}_n - y_n \right|^2 + \gamma \sum_{\forall i,k,l} \; \left| W_{{l-1},i,k}\right|^2    
\end{align}    
where $\gamma$ = $10^{-4}$ is the regularization coefficient and prevents the network from over-fitting to the noisy input data by avoiding unnecessarily large or active connections~\cite{Ref_DeepL}. The resulting regression function will be hence shaped by training the bias and weight parameters in the minimizing direction of $L(\cdot)$. In this regard, we follow the stochastic gradient descent via the well-known back propagation procedure, employing $1000$ epochs with the learning rate of $10^{-4}$ and batch size of 16. We denote the regression function obtained utilizing this model as $\varphi_{\text{ANN}}$ hereafter. Please see \cite{Ref_3, Ref_10} for more elaborations on the network implementation.

\subsection{Random forest regression (RFR)}
\begin{figure}[ht]   
	\begin{center}   
		\includegraphics[width=8cm, height=9cm]{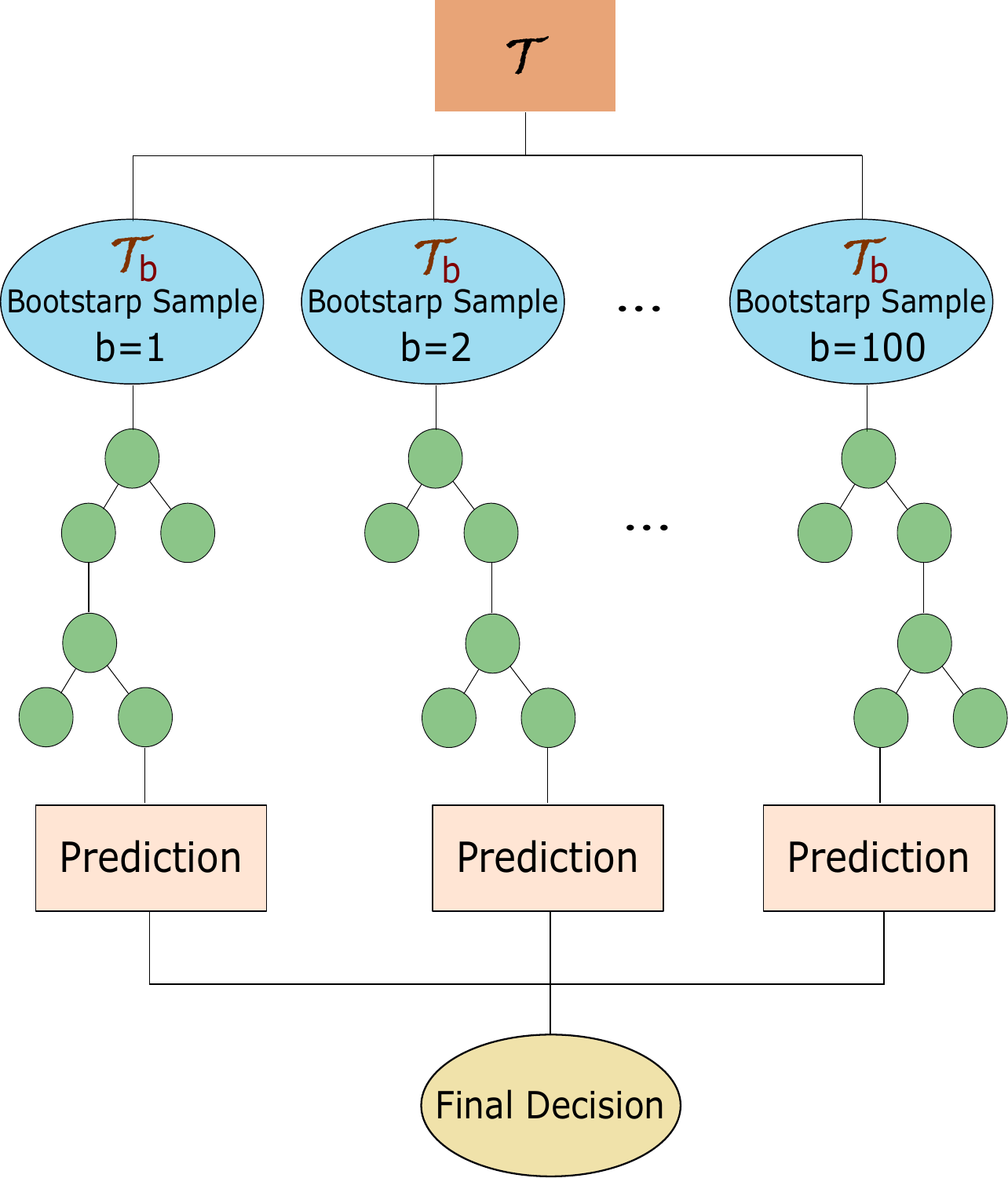}
		\caption{RFR diagram} \label{fig_DT}
	\end{center}
\end{figure}

The Random Forest Regression (RFR) has found applications in various engineering regression tasks due to its simplicity and good performance. In particular, it is possible to obtain reliable results with adjusting a few hyper parameters. As the name suggests, this algorithm generates a population of multiple Decision Trees (DT), each perform the role of a small tree-based regression task. The outcome of the RFR will be averaged value of the results of the individual decision DTs, which are implemented as explained in the following. Fig.~\ref{fig_DT} depicts the general procedure of the RFR method.
\subsubsection{Decision Tree (DT)}
To generate each DT, the RFR training algorithm applies the general technique of bootstrap aggregating, i.e., bagging, to tree learners. In particular, the data sets $\mathcal{T}_b, b \in \{1,\cdots,B\}$ are generated randomly \textit{with replacement} from the original data set $\mathcal{T}$. The intention is to shape the build several statistical representations of the data using the same data set. Each of the sets $\mathcal{T}_b$ are then used by an individual DT to calculate the regression task. 

In the training process of the $b$-th DT, the data set in $\mathcal{T}_b$ is partitioned in multiple stages, each time along a single chosen parameter. The choice of the partitioning parameter and the stopping criteria are given on the basis of the resulting \textit{information gain}, the reduction of the uncertainty as a result of making the partition, see \cite{Ref_14} for more elaborations.         

In this work, we have chosen the population of $B = 100$ and
the criteria to measure the quality of a split is chosen to be the MSE of the processed data at the DT, i.e., a potential split is chosen when obtaining a maximum variance reduction. Then nodes are expanded until all leaves are pure or until all leaves contain less than minimum samples split which is 2 and minimum samples leaf which is 1.
The RFR is implemented following the detailed steps reported in \cite[Subsection~2.1.1]{Ref_7}. We denote the regression function obtained by utilizing this model as $\varphi_{\text{RFR}}$ hereafter.

\subsection{Polynomial regression (PR)}

The polynomial regression (PR) extends the scope of classical linear regression methods by shaping a polynomial, and hence a nonlinear, regression function. Nevertheless, the training can be done similar to that of the linear regression training, by considering the higher order terms of the independent variables as a new independent variable in the training phase. In particular to our problem, the PR function can be formulated as
\begin{align}              
	\varphi_{\text{PR}} (\mathbf{x}) = \sum_{i=1}^{N} \sum_{j=0}^{M}     c_{ij} \left\{\mathbf{x}[i]\right\}^j \label{eq_PR}
\end{align}               
where $\mathbf{x}[i]$ denotes the $i$-th element of $\mathbf{x}$,~$N=8$ is the size of the independent parameter set in $\mathbf{x}$, $M = 3$ is the maximum polynomial order, $c_{ij}\in \mathbb{R}$ are the polynomial constants which shape the regression function. In order to obtain the intended polynomial function, we aim at finding the parameters $c_{ij}$ that lead to an accurate estimation. Please note that while the higher order terms in (\ref{eq_PR}) lead to a non-linear regression, they appear as a known value in the training phase, which correspond to a linear multivariable regression with extended dimensions. As a result, the optimum set of parameters $c_{ij}$ can be calculated similar to that of the linear multivariable regression using the least-squares method over the training set $\mathcal{T}$, please see \cite{Ref_15} for the details of the implemented PR method.  We denote the regression function obtained utilizing this model as $\varphi_{\text{PR}}$ hereafter.

\subsection{Multi-model regression} \label{MM_Regression}
We consider a weighted combination of the obtained solutions from the previously defined models, in order to improve the prediction accuracy. In particular, we define   
\begin{align} \label{eq_regreessionformula}                                          
	\varphi_{\text{MM}} := \beta_{\text{ANN}} \varphi_{\text{ANN}} + \beta_{\text{RFR}} \varphi_{\text{RFR}} + \beta_{\text{PR}} \varphi_{\text{PR}} 
\end{align}                                                         
as our mixed multi-model regression function, where $\beta_{\text{ANN}}, \beta_{\text{RFR}}, \beta_{\text{PR}} \in \mathbb{R}$ represent the weights associated with the individual regression models. It is obvious that the proposed weighted multi-model regression also subsumes, as its special case, each of the individual regression models by accordingly setting the weight to one or zero. The remaining question would be to choose weight in an optimum way, in order to improve the prediction. This is expressed via the optimization problem
\begin{align} 
	\underset{\beta_{\text{ANN}}, \beta_{\text{RFR}}, \beta_{\text{PR}} \in \mathbb{R}}{\text{minimize}}& \;\;\;\; \sum_{n \in \mathcal{T}} \; \left|  \varphi_{\text{MM}} (\mathbf{x}_n) - y_n \right|^2, \label{eq_3}\\
	\text{subjext~to} \;\;\;\;\; & \;\;\;\;\; \sum_{n \in \mathcal{T}}  \varphi_{\text{MM}} (\mathbf{x}_n) - \sum_{n \in \mathcal{T}} y_n  = 0, \label{eq_4}
\end{align}
where the constraint (\ref{eq_4}) is a popular regularization term used to enforce unbiasedness, which is a desirable estimator property for many applications~\cite{Ref_16}. Nevertheless, depending on the intended application, this constraint may be omitted in order to reduce MSE
\begin{align} 
	\underset{\beta_{\text{ANN}}, \beta_{\text{RFR}}, \beta_{\text{PR}} \in \mathbb{R}}{\text{minimize}}& \;\;\;\; \frac{1}{|\mathcal{T}|} \sum_{n \in \mathcal{T}} \; \left|  \varphi_{\text{MM}} (\mathbf{x}_n) - y_n \right|^2, \label{eq_5}
\end{align}
compromising the unbiasedness property, please see \cite{Ref_16} for more elaboration. It can be observed that both problems (\ref{eq_3})-(\ref{eq_4}) and (\ref{eq_5}) comply with the quadratic convex structure presented in \cite{Ref_17}. As a result, both problems can be solved efficiently via interior-point methods, employing the available numerical solvers, e.g., SeDuMi~\cite{Ref_solver1}~or Gurobi~\cite{Ref_solver2}.

\section{MODEL-BASED MULTI-OBJECTIVE MIXTURE DESIGN} \label{sec_optimize}
In this part, this is our intention to utilize the developed CS regression prediction model for the purpose of mixture optimization. Please note that a model-based optimization is of high significance to avoid the cost and delay associated with an experimental try-and-error approach, specially when considered for a multi-constraint, multi-objective design. In this regard, we employ the well-known NSGA-II algorithm~\cite{Ref_H} which is applied on the regression model developed in Section~\ref{sec_multimodel}, considering a multi-objective approach. In the following, we first define the corresponding multi-constraint, multi-objective optimization problem. Afterwards, we summarize the steps of the employed NSGA-II algorithm in order to efficiently solve the defined problem. The results of the proposed model-based optimization are given in Section~\ref{sec_ga_res}.

\subsection{Mixture Optimization: Problem Definition}
As mentioned previously, the dataset been used in this study contains $1030$ concrete mixture samples, each sample containing $8$ independent variables quantifying the age of curing ($C_{\text{AC}}$) with the unit of day, the content share of cement ($C_{\text{C}}$), content of blastfurnace slag ($C_{\text{BFS}}$), content of fly ash ($C_{\text{FLA}}$), content of water ($C_{\text{W}}$), content of fine aggregate ($C_{\text{FA}}$), content of coarse aggregate ($C_{\text{CA}}$) and content of superplasticizer ($C_{\text{SP}}$). The last $7$ independent variables, representing the content shares, have similar unit which is $kg/m^3$. The output (i.e., the dependent) variable, CS, is presented in MPa Unit. In relation to the choice of the mixture parameters, we take on two practical objectives. Firstly, it is of interest to satisfy a desired level of CS value. Moreover, it is of interest to minimize the resulting cost associated with the chosen mixture. Please note that these two constrains usually act in the opposite directions, e.g., an increase in the required CS value may necessitate the use of a more expensive mixture, which motivates the application of a multi-objective design framework. In this regard, the problem objectives (to be minimized) are defined as follows:
\begin{align}
	\text{\textbf{Objective~1:}}\;\;\;\; &\text{Cost}\left( C_{\text{AC}}, C_{\text{C}}, C_{\text{FLA}}, C_{\text{BFS}}, C_{\text{W}}, C_{\text{CA}}, C_{\text{FA}}, C_{\text{SP}}  \right)=\\ & \hspace{-25mm} \;  P_{\text{C}} C_{\text{C}} + P_{\text{FLA}} C_{\text{FLA}}+P_{\text{BFS}} C_{\text{BFS}} + P_{\text{W}} C_{\text{W}} + P_{\text{CA}} C_{\text{CA}} + P_{\text{FA}} C_{\text{FA}} + P_{\text{SP}} C_{\text{SP}},\nonumber  \\
	\text{\textbf{Objective~2:}}\;\;\;\; & \Big| \text{CS}_{\text{desired}} - \varphi_{\text{MM}} \left( C_{\text{AC}}, C_{\text{C}}, C_{\text{FLA}}, C_{\text{BFS}}, C_{\text{W}}, C_{\text{CA}}, C_{\text{FA}}, C_{\text{SP}} \right) \Big|,
\end{align} 
where the first and second objective respectively represent the cost and the difference of the predicted CS value via the developed regression model in Section~\ref{MM_Regression} with the desired (intended) CS value, associated with the mixture ($C_{\text{AC}},C_{\text{C}}, C_{\text{FLA}}, C_{\text{BFS}}, C_{\text{W}}, C_{\text{CA}}, C_{\text{FA}}, C_{\text{SP}}$). In the above expressions, the parameters $\big( P_{\text{C}},P_{\text{BFS}} ,P_{\text{FLA}}, P_{\text{W}}, P_{\text{CA}},$   $P_{\text{FA}}, P_{\text{SP}}  \big)$, respectively represent the unit prices of cement (0.11 $\$/kg$), blast-furnace slag (0.060 $\$/kg$), fly ash (0.055 $\$/kg$), water (0.00024 $\$/kg$), coarse aggregate (0.010 $\$/kg$), fine aggregate
(0.006 $\$/kg$) and superplasticizer (2.94 $\$/kg$).  

In addition to the defined problem objectives, the mixture parameters must satisfy some additional constraints in order to be practically feasible. For instance, the consumption of a particular element may be upper-bounded due to the limited availability or a physical limit. In particular, unit weight of each mixture element, as well as the constrains on range of the mixture parameters have been presented in detail in Tables~\ref{table_unt} and \ref{table_const}, respectively. The constraints indicating the practical range of each mixture element share is presented in Table~\ref{table_const}. The total unit volume constraint, collectively limiting the consumed elements in terms of the resulting volume is expressed as 
\begin{align}
	V_t &\left( C_{\text{C}}, C_{\text{FLA}}, C_{\text{BFS}}, C_{\text{W}}, C_{\text{CA}}, C_{\text{FA}}, C_{\text{SP}}  \right) = \\
	&\hspace{-3mm}C_{\text{C}}/\gamma_{\text{C}} + C_{\text{FLA}}/\gamma_{\text{FLA}} + C_{\text{BFS}}/\gamma_{\text{BFS}} + C_{\text{W}}/\gamma_{\text{W}} + C_{\text{CA}}/\gamma_{\text{CA}} + C_{\text{FA}}/\gamma_{\text{FA}} + C_{\text{SP}}/\gamma_{\text{SP}}\nonumber \\\nonumber & \nonumber\;\;\;\;\;\;\;\;\;\;\;\;\;\;\;\;\;\;\;\;\;\;\;\;\;\;\;\;\;\;\;\;\;\;\;\;\;\;\;\;\;\;= 1,
\end{align} 
where $\gamma_{\text{C}}, \gamma_{\text{FLA}}, \gamma_{\text{BFS}}, \gamma_{\text{W}}, \gamma_{\text{CA}}, \gamma_{\text{FA}}, \gamma_{\text{SP}}$ represent the unit weights of each mixture element, also see Table~\ref{table_unt} for detailed definitions. In the next subsection, we propose a procedure based on the NSGA-II algorithm in order to minimize the defined objective functions, while complying with the above-mentioned constraints on ratio, range and the resulting volume of the mixture parameters.

\begin{table}[h]
	\begin{center}
		\begin{tabular}{*{3}{c}}
			\hline
			\textbf{Component} & \textbf{{\scriptsize Unit name}} & \textbf{{\scriptsize \hspace{3mm}Unit weight (kg/m3)}}\\
			\hline
			\textbf{Cement ($\gamma_{\text{C}})$:} & {kg/m3}  & $3150$\\
			\textbf{Fly ash ($\gamma_{\text{FLA}})$):} & {kg/m3}  & $2500$\\
			\textbf{Blast-furnace slag ($\gamma_{\text{BFS}})$):} & {kg/m3}  & $2800$\\
			\textbf{Water ($\gamma_{\text{W}})$):} & {kg/m3}  & $1000$\\
			\textbf{Coarse aggregate ($\gamma_{\text{CA}})$):} & {kg/m3}  & $2500$\\
			\textbf{Fine aggregate ($\gamma_{\text{FA}})$):} & {kg/m3}  & $2650$\\
			\textbf{Superplasticizer ($\gamma_{\text{SP}})$):} & {kg/m3}  & $1350$\\
			\hline
		\end{tabular}
	\end{center}
	\caption{Unit weight of the input mixture parameters}
	\label{table_unt}
\end{table}

\begin{table}[h]
	\begin{center}
		\begin{tabular}{*{4}{c}}
			\hline
			\textbf{{\scriptsize Ratio}} & \textbf{{\scriptsize Expression}} & \textbf{{\scriptsize  \hspace{-4mm} Lower bound(\%)}}  & \textbf{{\scriptsize Upper bound(\%)}} \\
			\hline
			\textbf{{\scriptsize Cement/binder} } & {{\scriptsize $C_{C}$/($C_{C}$+$C_{BFS}$+$C_{FLA})$}}  & $26.4$  & $100$\\
			\textbf{{\scriptsize Water/binder}} & {{\scriptsize $C_{W}$/($C_{C}$+$C_{BFS}$+$C_{FLA})$}}  & $23.5$  & $90$\\
			\textbf{{\scriptsize Water/Cement}} & {{\scriptsize $C_{W}$/$C_{C}$}}   & $27$  & $188$\\
			\textbf{{\scriptsize Fly ash/binder}} & {{\scriptsize $C_{FLA}$/($C_{C}$+$C_{BFS}$+$C_{FLA})$}}  & $0$  & $55.2$\\
			\textbf{{\scriptsize Fly ash/Cement}} & {{\scriptsize $C_{FLA}$/$C_{C}$}}   & $0$  & $143$\\
			\textbf{{\scriptsize Blast-furnace/binder}} & {{\scriptsize $C_{BFS}$/($C_{C}$+$C_{BFS}$+$C_{FLA})$}}  & $0$  & $61$\\
			\textbf{{\scriptsize Blast-furnace/Cement}} & {{\scriptsize $C_{BFS}$/$C_{C}$}}  & $0$  & $158$\\
			\textbf{{\scriptsize Superpl./binder}} & {{\scriptsize $C_{SP}$/($C_{C}$+$C_{BFS}$+$C_{FLA})$}}   & $0$  & $5.6$\\
			\textbf{{\scriptsize Superpl./Cement}} & {{\scriptsize $C_{SP}$/$C_{C}$}}  & $0$  & $13$\\
			\textbf{{\scriptsize Coarse agg./total agg.}} & {{\scriptsize $C_{CA}$/($C_{CA}$+$C_{FA}$)}}  & $46.2$  & $65.2$\\
			\textbf{{\scriptsize Fine agg./total agg.}} & {{\scriptsize $C_{FA}$/($C_{CA}$+$C_{FA}$)}}  & $34.8$  & $53.8$\\
			\textbf{{\scriptsize Coarse agg./binder}} & {{\scriptsize $C_{CA}$/($C_{C}$+$C_{BFS}$+$C_{FLA})$}}  & $118$  & $563$\\
			\textbf{{\scriptsize Fine agg./binder}} & {{\scriptsize $C_{FA}$/($C_{C}$+$C_{BFS}$+$C_{FLA})$}}   & $106$  & $423$\\
			\textbf{{\scriptsize Binder/total weight}} & {{\scriptsize ($C_{C}$+$C_{BFS}$+$C_{FLA})$)/$\sum(C_{All})$} }   & $8.5$  & $27$\\
			\textbf{{\scriptsize Aggregate/total weight}} & {{\scriptsize ($C_{CA}$+$C_{FA}$)/$\sum(C_{All})$} }  & $64$  & $84$\\
			\hline
		\end{tabular}
	\end{center}
	\caption{Ratio and Range constraints for the mixture parameters}
	\label{table_const}
\end{table}

\begin{figure}[!ht]
	\begin{center}
		\includegraphics[width=11cm]{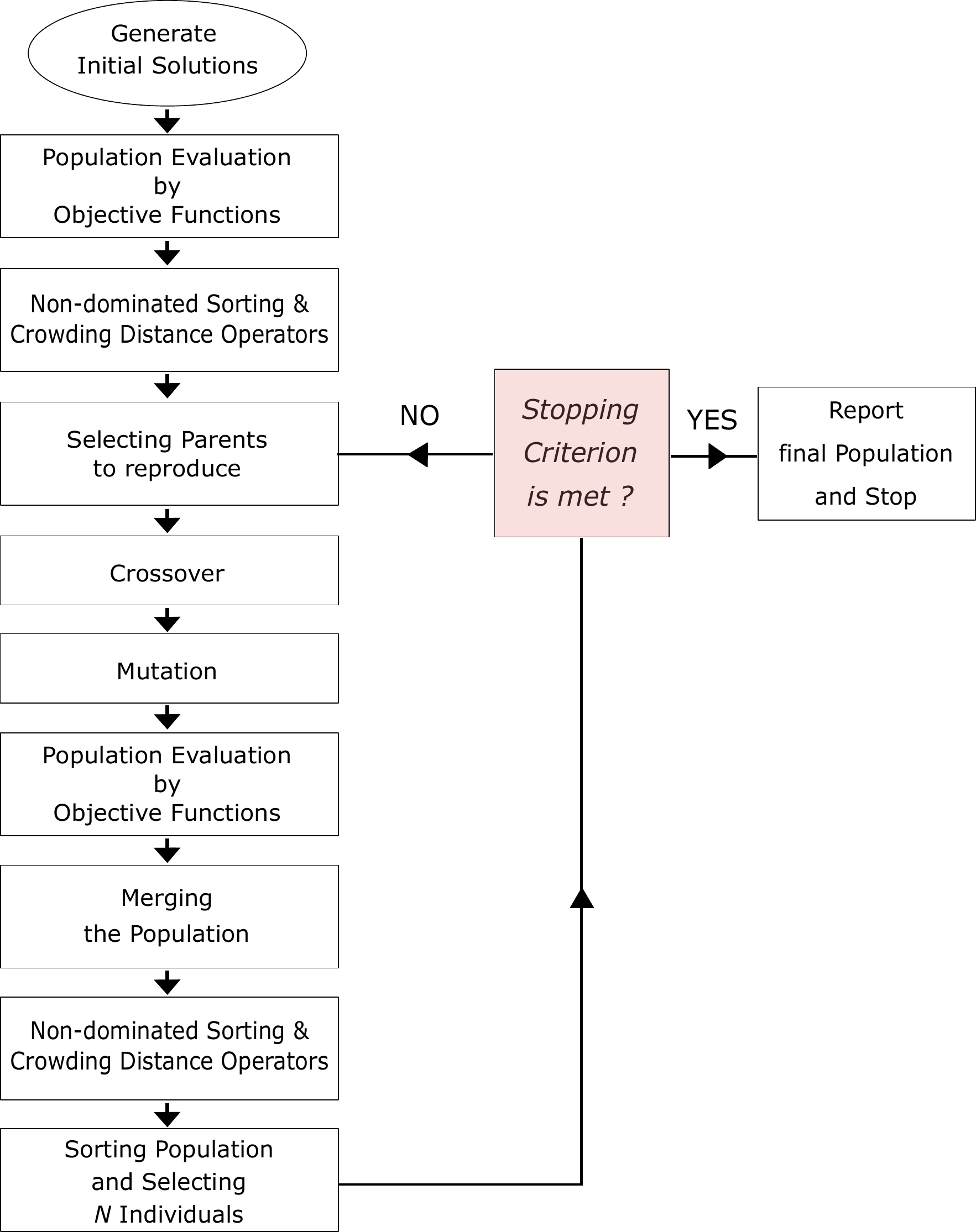}
		\caption{Flowchart of NSGA-II algorithm} \label{fig_gaFlow}
	\end{center}
\end{figure} 
\subsection{NSGA-II: Algorithm Procedure}
Genetic algorithm (GA) is a well-known heuristic algorithm based on Darwin’s natural selection theory that belongs to the larger class of evolutionary algorithms (EA)~\cite{Ref_I}. The classic GA seeks to maximize an objective function, under a pre-given set of constraints, by sequentially generating population of the parameter sets and refine the remaining individuals towards the selected objective. Nevertheless, the algorithm does not consider the possibility of a multi-objective and multi-constraint framework, which is the interest of this paper. In this regard, we employ an extension of the GA, namely, NSGA-II method, which is adjusted to tackle the desired multi-objective framework. In the following, we describe the implemented steps to solve the defined problem. 
\subsubsection{Random population generation}  At the beginning of GA implementation, we generate the first population in a random call which constitute the first-generation set of chromosomes. Each chromosome correspond to a vector of scalar values, containing the quantities of the mixture parameters. This step is confined by the general constraints that we define at the first place. For example: the minimum and maximum range for any parameters that contribute in forming a chromosome, and the proportion between them that needs to be satisfied.
\subsubsection{Sorting}
NSGA-II is highly similar to the traditional genetic algorithm with a difference in ‘sorting’ that is the next step in GA implementation. For sorting the chromosomes by their suitability in, two steps are required: \textit{Non dominated sort} and \textit{Crowd distance calculation}:
\textit{Non dominated sort}: Sorting process based on non-domination criteria of the population that has been initialized. In this case a Gene with a smaller or equal value of the resulting price and less value regrading the difference of the target CS and sample CS is said to be dominant to another sample. An attribute as rank is then assigned to each chromosome as the number of times that the same chromosome is dominated by others. The value rank will hence be an initial factor for the sorting mechanism.

\textit{Crowd distance calculation}: Once the sorting is complete, the crowding distance value is assigned. The aforementioned distance corresponds to the distance of a specific chromosome to the nearest one, hence indicating how much the presence of the specific chromosome is informative to the collection of the chromosomes.
\subsubsection{Selection}
The next step of implementation is ‘selection’. The individuals in population are selected based on their rank in ‘Non dominated sort’ and their distance value in ‘crowding distance’.

\subsubsection{Cross-Over and Mutation}
The two other steps in GA and also the NSGA-II are crossover and mutation. In each iteration or generation, these operators are used on a population of all possible solutions, in order to develop their fitness and expand the associated search space. For a schematic of the algorithm procedure please see~Fig.~\ref{fig_gaFlow}. Moreover, for a detailed discussion on the aforementioned steps and the reasoning behind them please see~\cite{Ref_H}.

In this study, our initial population consist of $200$ genes which are generated at random and comply with the previously defined constraints, e.g., range constrain for the parameters. Furthermore, the ‘roulette wheel’ method is used in order to well-distribute the chance of being chosen by competency. In the other words, the selection of a gene is done at random, with a probability which is proportional to its fitness, e.g., the rank and/or distance values. Also in each iteration $100$ children genes are generated and $50$ best samples are selected in accordance to their ranks and crowding distances. The performance of the proposed scheme is studied in the next section via numerical simulations.

\section{NUMERICAL EVALUATION} \label{sec_results}
In this part, we evaluate the accuracy of the implemented regression methods for predicting the concrete CS, given the input mixture parameters. In order to train and evaluate the regression methods, we utilize the mixture data set collected in \cite{Ref_2}, where the data set is divided into the training and the test data sets, i.e., $\mathcal{T}, \mathcal{T}^o$ as explained in Section~\ref{sec_data}. In particular, we evaluate the prediction mean squared error over the training and test data sets, employing the regression functions $\varphi_{\text{ANN}}, \varphi_{\text{RFR}}, \varphi_{\text{PR}}$, as well as $\varphi_{\text{MM}}$, which represents the proposed multi-model regression with optimized weights. 

\subsection{Single-model evaluation}

\begin{figure}[ht]
	\begin{center}
		\includegraphics[width=12cm]{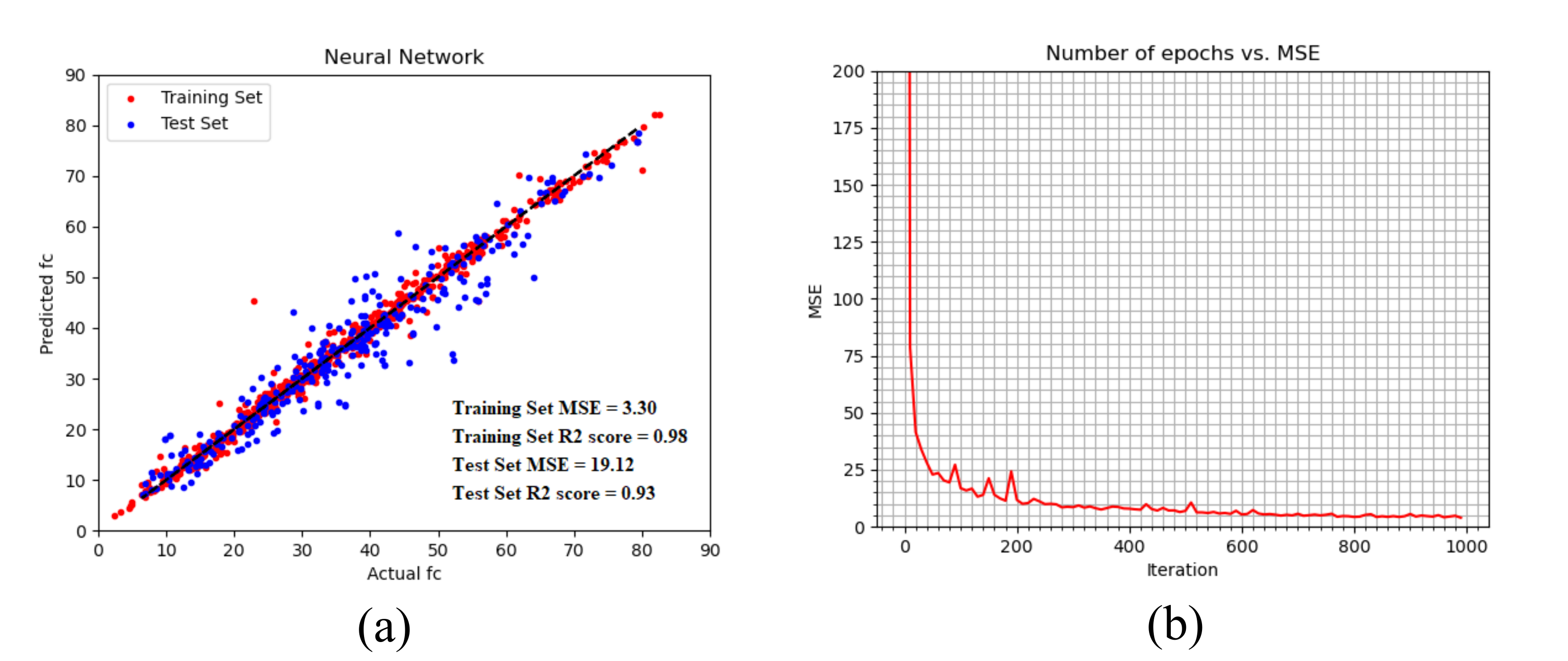}
		\caption{(a) Resulting prediction instances for ANN: Training set vs. Test set, (b) ANN MSE reduction vs. epochs} \label{fig_sim_ANN}
	\end{center}
\end{figure}

In Fig.~\ref{fig_sim_ANN} the MSE of the prediction behavior of the ANN is depicted, both after and during the training phase. In Fig.~\ref{fig_sim_ANN}a, the resulting MSE and $R^2$ score over the training and test data sets are depicted, utilizing the trained regression function $\varphi_{\text{ANN}}$. Please note that the small MSE value over the training set represents how well the regression function has been able to adapt to the nature of the input data, hence, learning the input-output relations over the training set. Nevertheless, the  MSE value is then evaluated over the test set to indicate the generalization capability, i.e., how well the regression function is capable of providing an accurate prediction for the input data which does not belong to the training set. In Fig.~\ref{fig_sim_ANN}b, the MSE of the ANN output is depicted over the training data set, for different number of epochs. Due to the employed backpropagation-based stochastic gradient descent, the value of the MSE converges stochastically to a minimum. It is observed that the employed $1000$ epochs is sufficient to obtain a stable ANN performance.    

\begin{figure}[ht]
	\begin{center}
		\includegraphics[width=12cm]{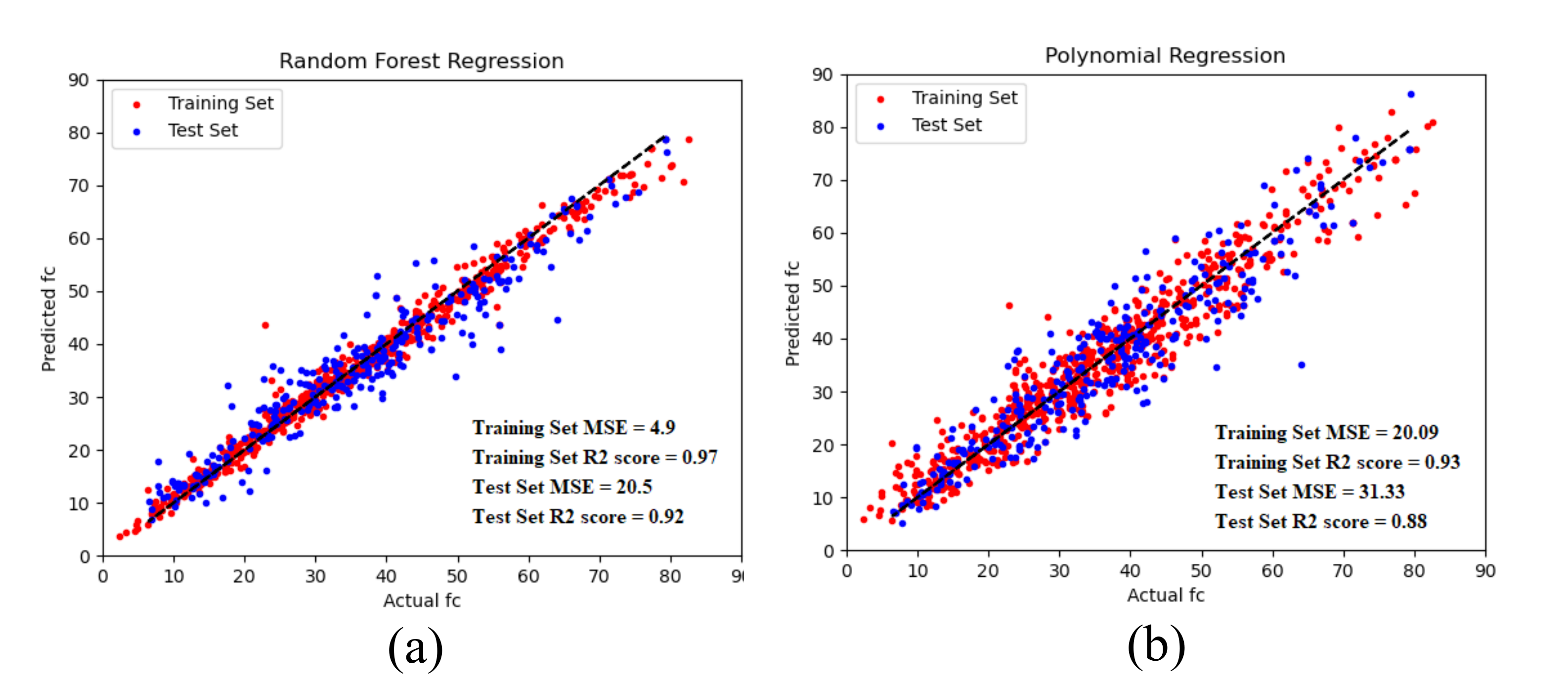}
		\caption{Resulting prediction instances: Training set vs. Test set, (a): RFR, (b): PR} \label{fig_sim_RFR_PR}
	\end{center}
\end{figure}

Please note that the multi-layered ANN regression is known to be the best-performing model among the standard regression methods for several investigated problems, see \cite{Ref_7, Ref_12}. Nevertheless, this is the intention of the present work to evaluate the accuracy improvement that can be obtained by employing a multi-model regression, compared to the usage of the single best model. In this regard, the prediction accuracy of the other regression methods, i.e., $\varphi_{\text{RFR}}, \varphi_{\text{PR}}$ have been evaluated in Fig.~\ref{fig_sim_RFR_PR}. It is observed that the RFR achieves a higher prediction accuracy compared to that of the PR, both over the training and the test data sets. Nevertheless, it is observed that both methods have been outperformed by the ANN benchmark, both over the test and the training data sets.
\subsection{Multi-model evaluation}
\begin{figure}[!ht]
	\begin{center}
		\includegraphics[width=12cm]{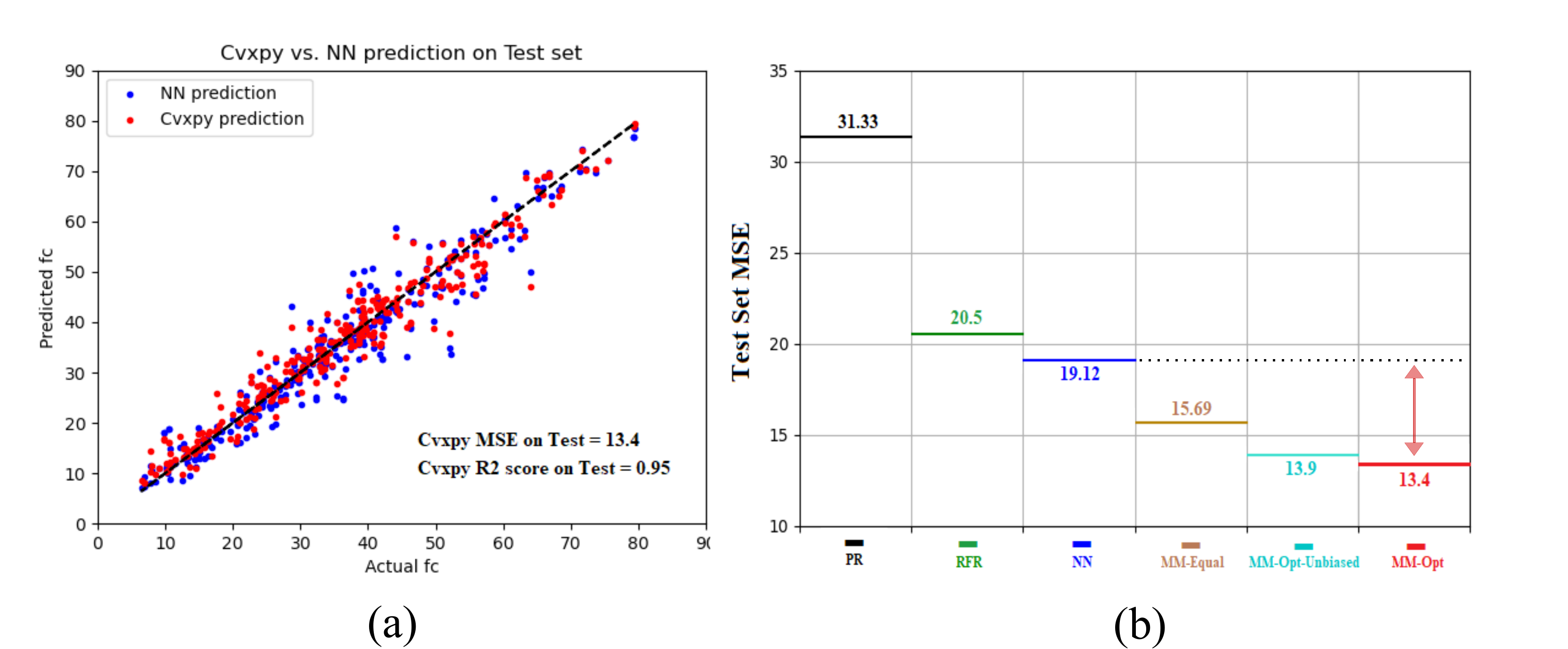}
		\caption{(a) Resulting prediction instances on Test set: ANN vs. MM-Opt, (b) comparison of the resulting MSE, using different regression methods} \label{fig_sim_cvx}
	\end{center}
\end{figure}

In Fig.~\ref{fig_sim_cvx}a, the prediction accuracy of the optimized multi-model regression, proposed in Section~\ref{MM_Regression} is reported compared to the best single-model regression, i.e., the ANN model. it is observed that the multi-model regression reaches a significantly higher prediction accuracy over the test data set, both in terms of the prediction MSE as well as the obtained $R^2$ score. The obtained model accuracy gains are depicted in Fig.~\ref{fig_sim_cvx}b, where the optimized multi-model regression achieves a significantly lower prediction MSE, both compared to the optimized best single-model regression, i.e., the trained ANN model, as well as the non-optimized multi-model regression, i.e., the proposed regression model in Section~\ref{MM_Regression} without weight optimization. We should note that the achieved accuracy gain is obtained at the cost of an additional processing overhead for multi-model implementation, which is negligible due to the increasing memory and processing power of the state-of-the-art processors. 
\begin{table}[!ht] 
	\begin{center}
		\begin{tabular}{*{4}{c}}
			\hline
			\textbf{{\hspace{-2mm}\scriptsize Target CS ${\tiny  (MPa)}$}} & \textbf{{\hspace{2.5mm}\scriptsize Sample CS ${\tiny  (MPa)}$}} & \textbf{{\hspace{2.5mm}\scriptsize Distance ${\tiny  (MPa)}$}}  & \textbf{{\hspace{2.4mm}\scriptsize Cost ${\tiny  (\$)}$}} \\
			\hline
			\textbf{} $20$   & $18.82$ & $1.18$  & $46.32$\\
			\textbf{} $25$   & $24.37$ & $0.63$  & $47.38$\\
			\textbf{} $30$   & $28.63$ & $1.37$  & $47.89$\\
			\textbf{} $35$   & $36.67$ & $1.67$  & $48.37$\\
			\textbf{} $40$   & $40.47$ & $0.47$  & $49$\\
			\textbf{} $50$   & $48.12$ & $1.88$  & $51.05$\\
			\textbf{} $60$   & $58.03$ & $1.97$  & $56.15$\\
			\textbf{} $70$   & $68.1$  & $1.9$   & $63.79$\\
			\textbf{} $75$   & $76$    & $1$     & $66.4$\\
			\hline
		\end{tabular}
	\end{center}
	\caption{The target CS values and the resulting cost obtained from the NSGA-II algorithm.}
	\label{table_res}
\end{table}

\subsection{Model-based mixture optimization}\label{sec_ga_res}
\begin{figure}[!h]
	\begin{center}
		\includegraphics[width=11cm]{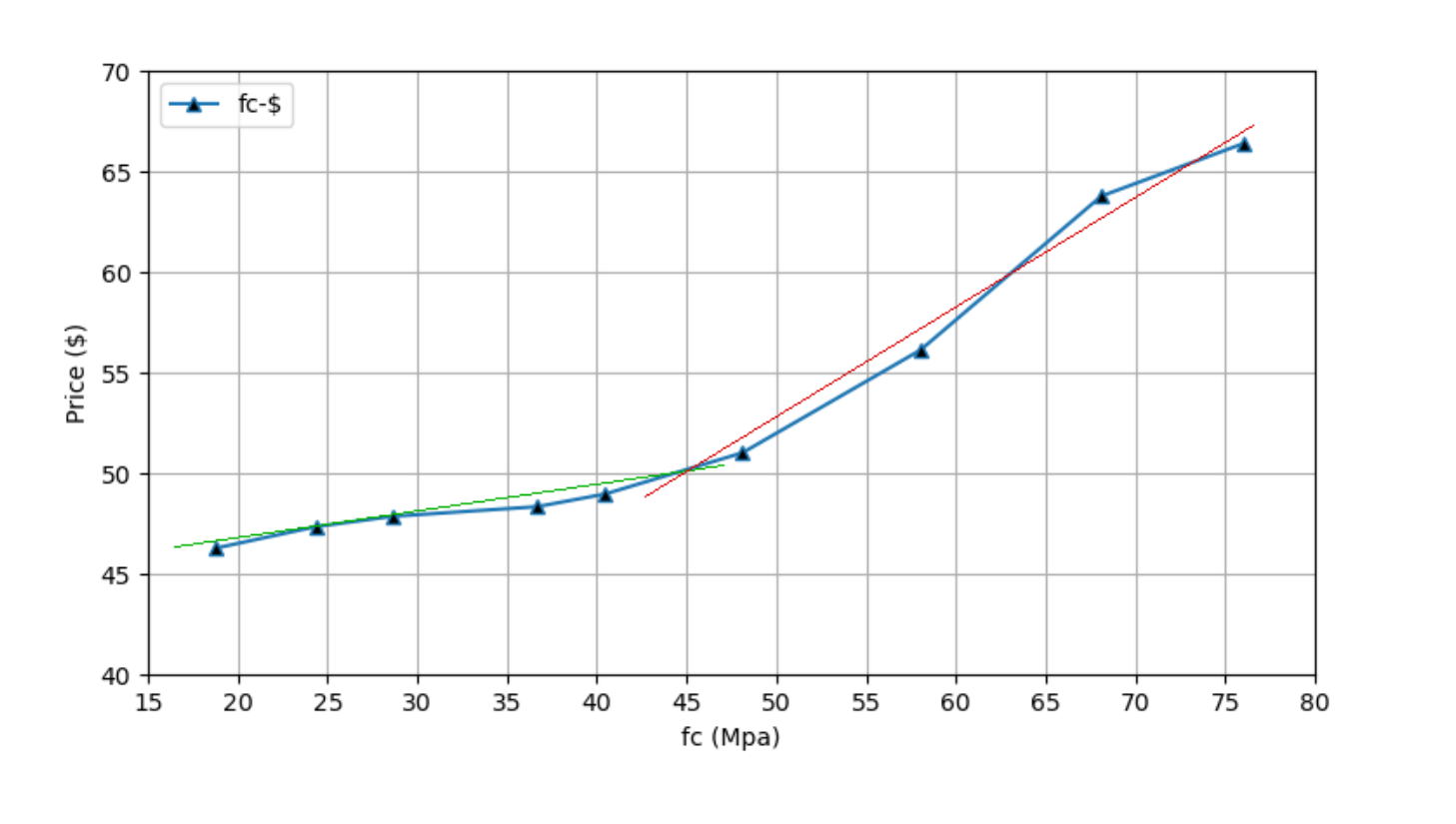}
		\caption{Variation of the resulting cost w.r.t. the required target CS. The slope of cost increases for a higher target CS region.} \label{fig_res}
	\end{center}
\end{figure}

By employing the NSGA-II-based procedure introduced in Section~\ref{sec_optimize} and employing the multi-model prediction to examine the objective values associated with each gene, we have numerically obtained the Pareto front representing the optimal cost and CS strength trade-off, i.e., in order to find closest sample to desired CS with lowest price complying with all the stated problem constraints. In this regard, we have obtained $9$ optimum samples (with age of curing of 28 days) and different desired CS which are depicted in Table~\ref{table_res}. 
Considering the result and cost associated with the respective CS values, with a relatively small change in the mixture parameters and a modest increase in the resulting cost, the obtained CS can be increased from 20 to 50 MPa, with a minimal additional costs. However, it is observed that in order to achieve a higher levels of the CS, e.g., in the range of $60-80$ MPa, a significantly higher increase in cost is imposed. Fig.~\ref{fig_res} depicts the aforementioned trend regarding the increase in the total costs. It is observed from Fig.~\ref{fig_res} that in average, for the first range of the cost-CS curve, an increase in the CS in the value of 1 MPa necessitates approximately $14\%$ increase in cost (see the slope of the green curve). Nevertheless, for a higher range of the studied CS, e.g., the second range, this ratio increases to approximately $41\%$, please see the slope of the red curve. The latter observation confirms an earlier statement that the increase in the cost will be higher when moving towards a higher CS region. 

\section{CONCLUSION} \label{sec_conclude}
In this work, the problem of model-based prediction and heuristic optimization of CS level for a known concrete mixture is studied. We have proposed a convex optimization-based framework for implementing and combining multiple regression models, with the goal of achieving a higher prediction accuracy. We have shown that the optimized multi-model regression achieves a significantly higher prediction accuracy, at the cost of a slightly higher complexity associated with the concurrent implementation of multiple regression models.
Moreover, employing the proposed NSGA-II-based algorithm, a Pareto front of the cost-CS trade-off has been obtained, including $9$ optimized samples with different CS values. Please note that the achieved improvement in the prediction accuracy is promising, considering the relatively limited available data set for a specific concrete mixture setup and the expensive experimental procedures, which motivates the implementation of more accurate models for the purpose of \textbf{\textit{$f_c$}} prediction and optimization.



\end{document}